\documentclass[a4paper, 10pt, conference]{ieeeconf}     
\usepackage{FG2025}

\FGfinalcopy % *** Uncomment this line for the final submission

\IEEEoverridecommandlockouts                              
\usepackage{graphics} % for pdf, bitmapped graphics files
\usepackage{epsfig} % for postscript graphics files
\usepackage{mathptmx} % assumes new font selection scheme installed
\usepackage{times} % assumes new font selection scheme installed
\usepackage{amsmath} % assumes amsmath package installed
\usepackage{amssymb}  % assumes amsmath package installed
\usepackage{float}
\usepackage{multicol}
\usepackage{subcaption}
\usepackage{textcomp}
\usepackage{url}

\usepackage{xcolor}
\usepackage[colorlinks]{hyperref}
\hypersetup{colorlinks=true, linkcolor=blue, urlcolor=red, citecolor=red}

\def\FGPaperID{128} % *** Enter the FG2025 Paper ID here

\title{\LARGE \bf Towards Iris Presentation Attack Detection with Foundation Models}

\author{\parbox{16cm}{\centering
    {\large Juan E. Tapia$^1$ and Lázaro Janier González-Soler$^1$, and Christoph Busch$^1$}\\
    {\normalsize
    $^1$ da/sec-Biometrics and Internet Security Research Group, Darmstadt, Germany}}\\
    e-mails:juan.tapia-farias, lazaro-janier.gonzalez-soler and christoph-busch{(@h-da.de)}
    \thanks{This work is supported by the European Union’s Horizon 2020 research and innovation program under grant agreement 101121280 (EINSTEIN) and by the German Federal Ministry of Education and Research and the Hessian Ministry of Higher Education, Research, Science and the Arts within their joint support of the National Research Center for Applied Cybersecurity ATHENE.}% <-this % stops a space
}

\begin{document}

\ifFGfinal
\thispagestyle{empty}
\pagestyle{empty}
\else
\author{Anonymous FG2025 submission\\ Paper ID \FGPaperID \\}
\pagestyle{plain}
\fi
\maketitle

%%%%%%%%%%%%%%%%%%%%%%%%%%%%%%%%%%%%%%%%%%%%%%%%%%%%%%%%%%%%%%%%%%%%%%%%%%%%%%%%
\begin{abstract}
Foundation models are becoming increasingly popular due to their strong generalization capabilities resulting from being trained on huge datasets. These generalization capabilities are attractive in areas such as NIR Iris Presentation Attack Detection (PAD), in which databases are limited in the number of subjects and diversity of attack instruments, and there is no correspondence between the bona fide and attack images because, most of the time, they do not belong to the same subjects. This work explores an iris PAD approach based on two foundation models, DinoV2 and VisualOpenClip. The results show that fine-tuning prediction with a small neural network as head overpasses the state-of-the-art performance based on deep learning approaches. However, systems trained from scratch have still reached better results if bona fide and attack images are available. 
\end{abstract}

%%%%%%%%%%%%%%%%%%%%%%%%%%%%%%%%%%%%%%%%%%%%%%%%%%%%%%%%%%%%%%%%%%%%%%%%%%%%%%%%
\section{Introduction}
During the last few years, several efforts have been developed to create an Iris Presentation Attack Detection (PAD) system based on NIR iris images \cite{Iris_survey} and deep learning approaches \cite{Hybrid-tapia}. However, according to the results reported in the state of the art, this topic is still an open challenge \cite{Livdet2023}. One of the most relevant factors is the available dataset, which presents a reduced number of images and subjects compared to other biometrics modalities such as face and fingerprint.

While deep learning has transformed computer vision tasks, current approaches have several significant challenges. First, typical vision datasets are labour-intensive and expensive to create, yet they often cover only a limited set of visual concepts. Second, standard vision models are effective at one task but require significant effort to adapt to new ones. Finally, models that perform well on benchmarks frequently show disappointing results in cross-dataset testing, raising concerns about the generalization capabilities of computer vision to be applicable to different tasks. 

On the other hand, today, one of the challenges is to obtain a large number of images (bona fide and attacks) based on privacy concerns for the training method based on deep learning and vision transformers approaches. The generation of synthetic images has been studied in order to complement the training process and obtain robust iris PAD algorithms with generalisation capabilities in a cross-dataset scenario \cite{jose-iris}. Most iris datasets contain a low number of subjects with many images per subject.

With the Large Languages Models (LLM), a new focus of the analysis has been developed, creating downstream tasks focusing on text, images, or a combination of both. For the computer vision task, two interesting approaches based on the image prediction capabilities, such as DinoV2 \cite{dinov2} and VisualOpenClip \cite{cherti2023reproducible, Radford2021LearningTV,  schuhmann2022laionb}. These two networks were trained with a self-supervised process.

DinoV2 foundation model generates universal features suitable for both image-level visual tasks, such as image classification and video understanding, as well as pixel-level visual tasks, including depth estimation and semantic segmentation. DinoV2 was trained using 140 million images without labels.

The Contrastive Language-Image Pre-Training (CLIP) learns visual concepts from natural language supervision and can be applied to any visual classification by providing the names of the visual categories to be recognized. The Clip model was trained with 400 million image and text pairs. 

Building on these previous approaches, this work emphasizes the considerable potential of foundation models across various tasks. For the first time, it explores the impact of fine-tuning classification on iris presentation attack detection, revealing a significant improvement in the results. 

The main contribution of our work is to demonstrate that Iris PAD based on the foundation model is feasible using only a small neural network as a head and outperforming all the traditional methods based on deep learning.

The rest of the paper is organized as follows: Section \ref{sec:related} reviews the related works. Section \ref{sec:method} explains the method, metrics and dataset. Section \ref{sec:expandresults} explains the experiments and results obtained, and Section \ref{sec:conclusions} draws the conclusions of this work.

\section{Related work}
\label{sec:related} 

A good starting point related to the method and datasets proposed is the last two competitions: Liveness Presentation Attack Detection on Iris Images hosted by The International Joint Conference in Biometrics (IJCB) in versions 2020 (LivDet2020)\cite{Livdet2020} and 2023 (LivDet2023) \cite{Livdet2023}. Both competition results have shown that a relevant gap exists based on the number of images (bona fide and attacks) used and the potential influence of iris synthetic images. One essential point to highlight is whether the bona fide subject in the dataset has any correlation with the attack images. While this can not be granted, this task is even more challenging.

The LivDet2020 reported results on bona fide and attacks such as Printed Irises, Patterned Contact Lenses, Fake/Prosthetic/Printed Eyes, Eyes Displayed on Kindle, and Cadaver Irises. The test set was sequestered with no access to teams. The winning team's approach achieved the lowest BPCER of 0.46\% among all nine algorithms across three categories. This result aligns well with the operational goal of Presentation Attack Detection (PAD) algorithms: to accurately identify genuine presentations (i.e., minimizing the system's False Rejection Rate) while maximizing the detection of attacks.

The LivDet2023 reported results on bona fide and attacks such as Paper Print-out irises, Cosmetic contact lenses, Cosmetic contact lenses on the printed eye,  Eye dome on the printed eye, Doll eye, Electronics display, Cosmetic contact lenses on the doll eye, Synthetic irises in low, medium and high-quality. The winning team achieved a BPCER of 35.40\%, with an APCER of 37.31\%, which weights all PAI categories equally.

The state of the art also shows the relevance of incorporating synthetic images in the training process because of the number of relevant attacks in the dataset representing a real threat. Previous work had developed a new method to generate synthetic iris images based on a Generative Adversarial Network (GAN) focusing on PAD \cite{boyd, jose-iris, Iris_survey} and the identity preservation \cite{Spoof-iris}.

Recent studies have begun to explore the potential of foundation models in biometrics, but there is still significant room for improvement in their applications. Researchers introduced the term "foundation model" to describe machine learning models that are trained on a wide range of generalized and unlabeled data. These models are capable of performing various general tasks, including understanding language, generating text and images, and engaging in natural language conversations. 

One notable example is Iris-SAM \cite{iris-sam}. This model has been fine-tuned on ocular images for iris segmentation and has achieved an average segmentation accuracy that exceeds the best baseline by a considerable 10\% on the ND-IRIS-0405 dataset.

In \cite{iris_chatgpt} the researchers extensively evaluated Chat-GPT-4 ability to compare and analyse iris patterns and compared its performance against established iris recognition systems. This assessment aimed to determine the potential use of iris biometric recognition.

Another example is Arc2Face \cite{arc2face}, an identity-conditioned foundation model that generates photorealistic images based on a person’s ArcFace embedding. To demonstrate the performance of the generated data, the researchers trained a face recognition model on synthetic images produced by their model, achieving better results than those obtained from existing synthetic datasets.

Additionally, Chettaoui et al. \cite{chettaoui}. Investigate the capabilities of foundation models in face recognition. They propose fine-tuning these models using LOw-rank adaptation RAnk (LoRA), which allows for modifying the weights of the network while keeping all other layers of the foundation model frozen.

According to our state-of-the-art analysis, no previous work has analysed the impact of foundation models in Iris PAD, which focalize and motivate our approach.

%%%

\begin{figure*}[!htb]
\centering
    \subcaptionbox{\centering Bona fide}{\includegraphics[scale=0.19]{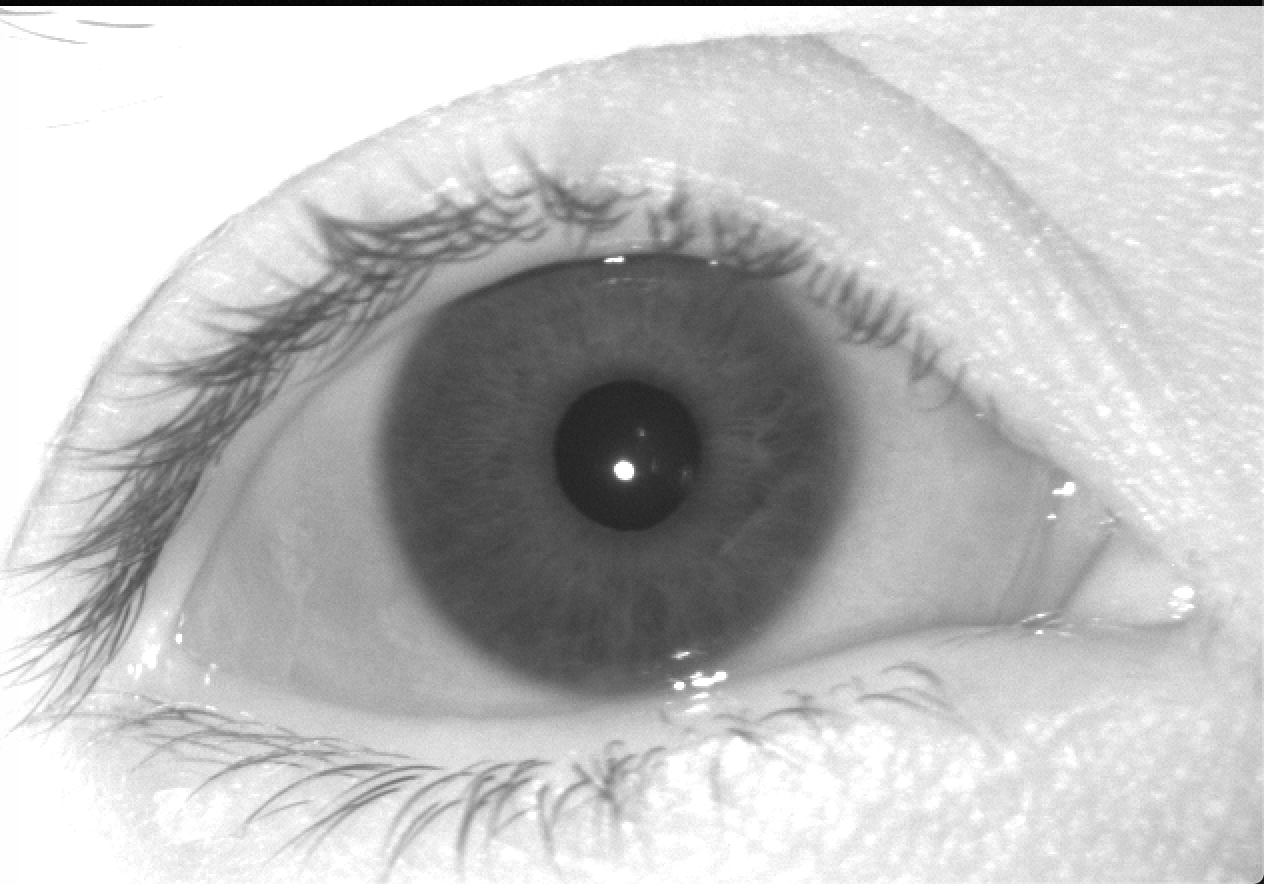}}
    \hfill
    \subcaptionbox{\centering Cadaver (post-mortem subject) eye}{\includegraphics[scale=0.18]{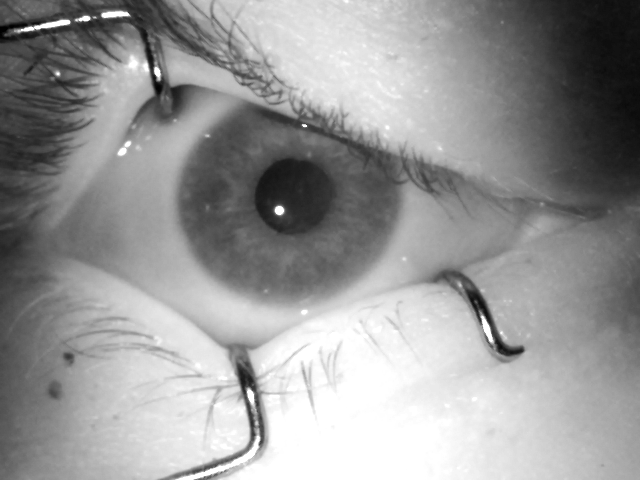}}
    \hfill
    \subcaptionbox{\centering Cosmetic contact lens}{\includegraphics[scale=0.20]{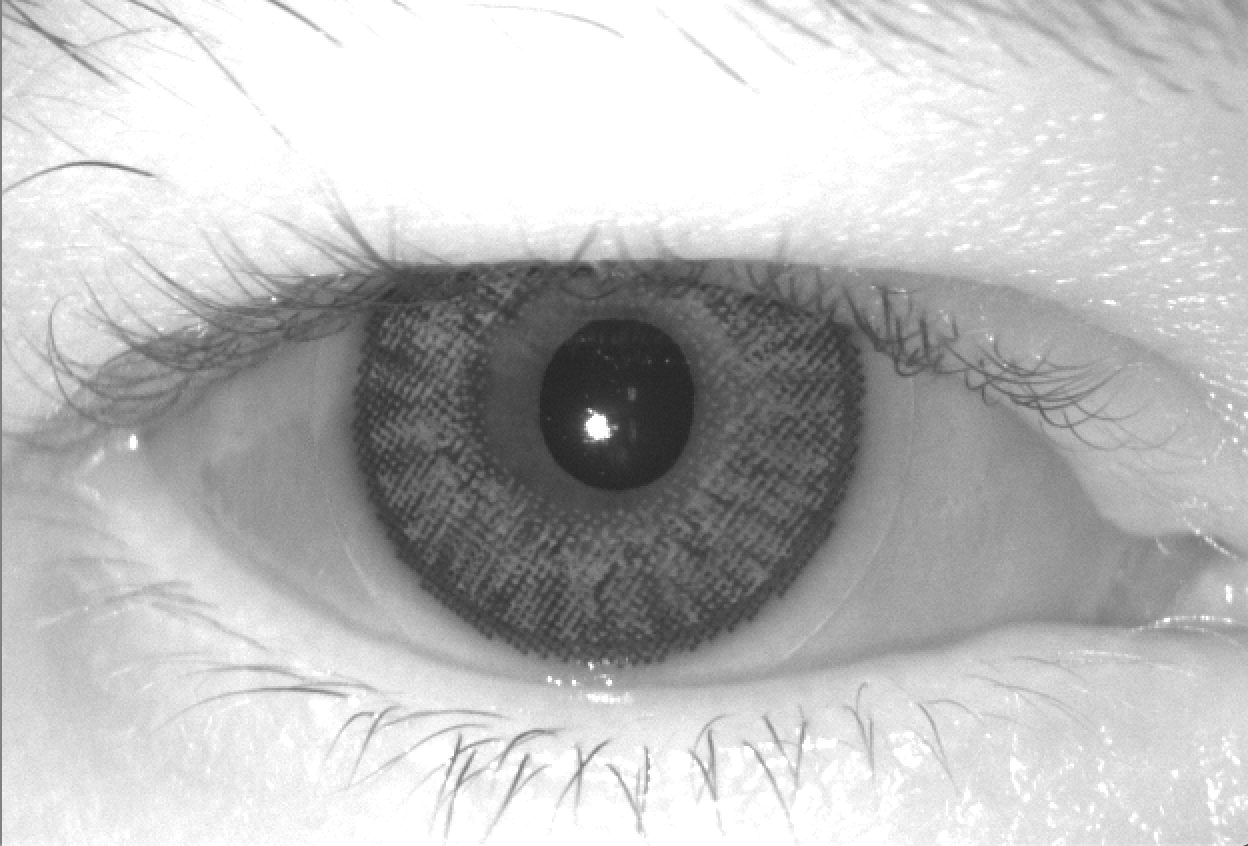}}
    \hfill
    \subcaptionbox{\centering Electronic display}{\includegraphics[scale=0.18]{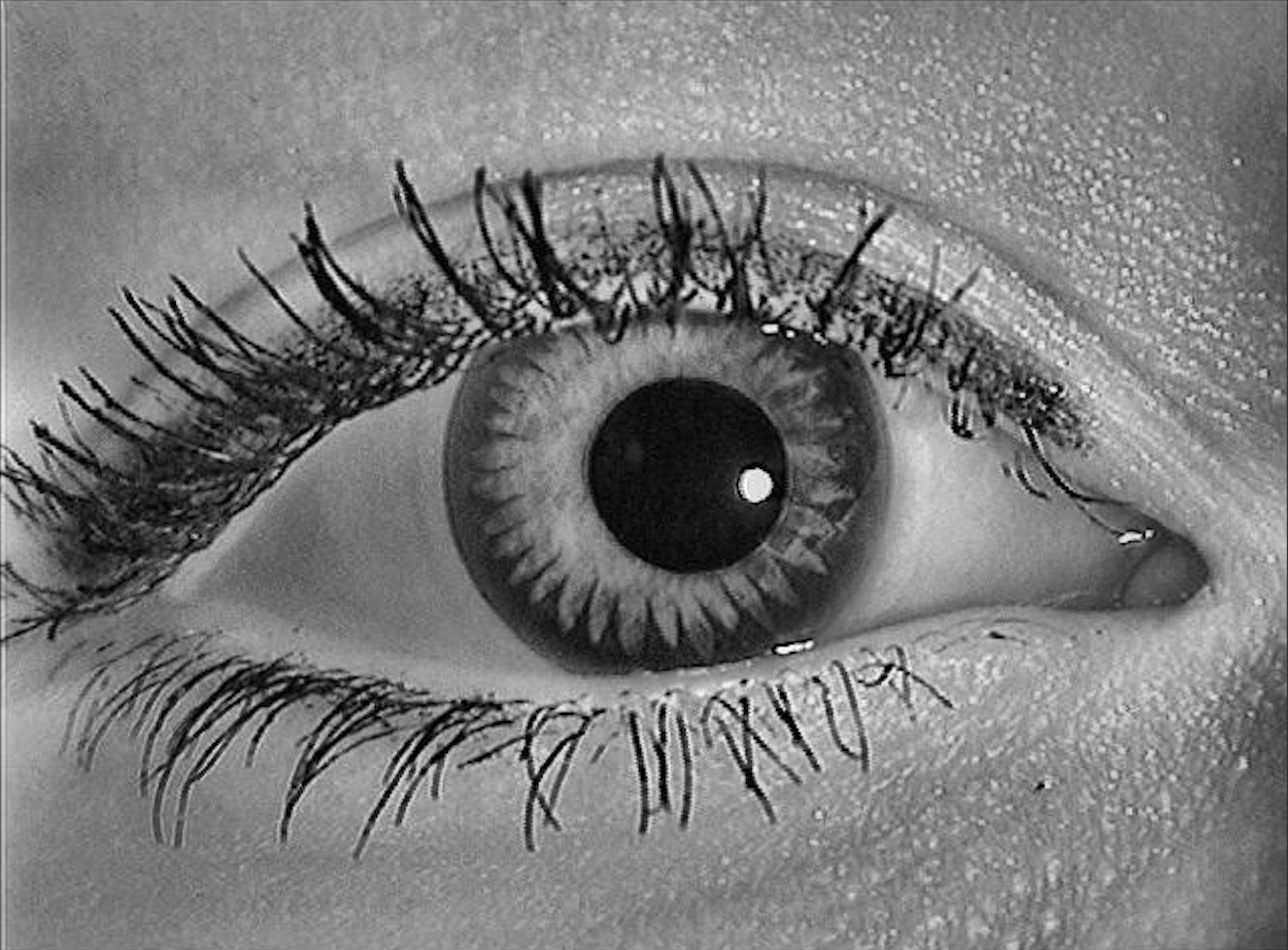}}
    \hfill
    \subcaptionbox{\centering Print-out LivDet-Iris 2020}{\includegraphics[scale=0.18]{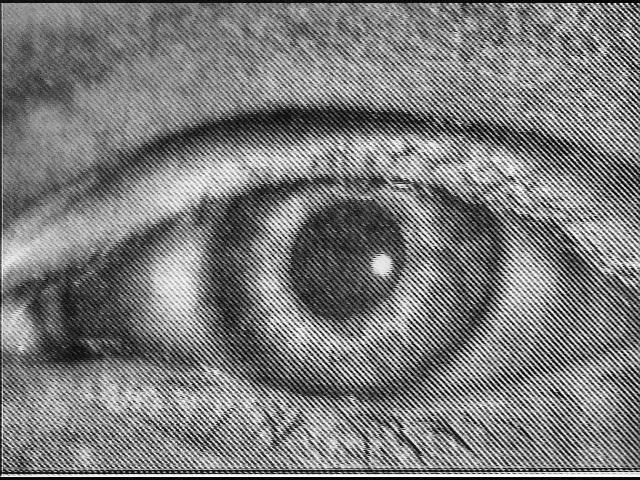}}
    \hfill 
    \subcaptionbox{\centering Print-out Motorola MotoG4Play}{\includegraphics[scale=0.09]{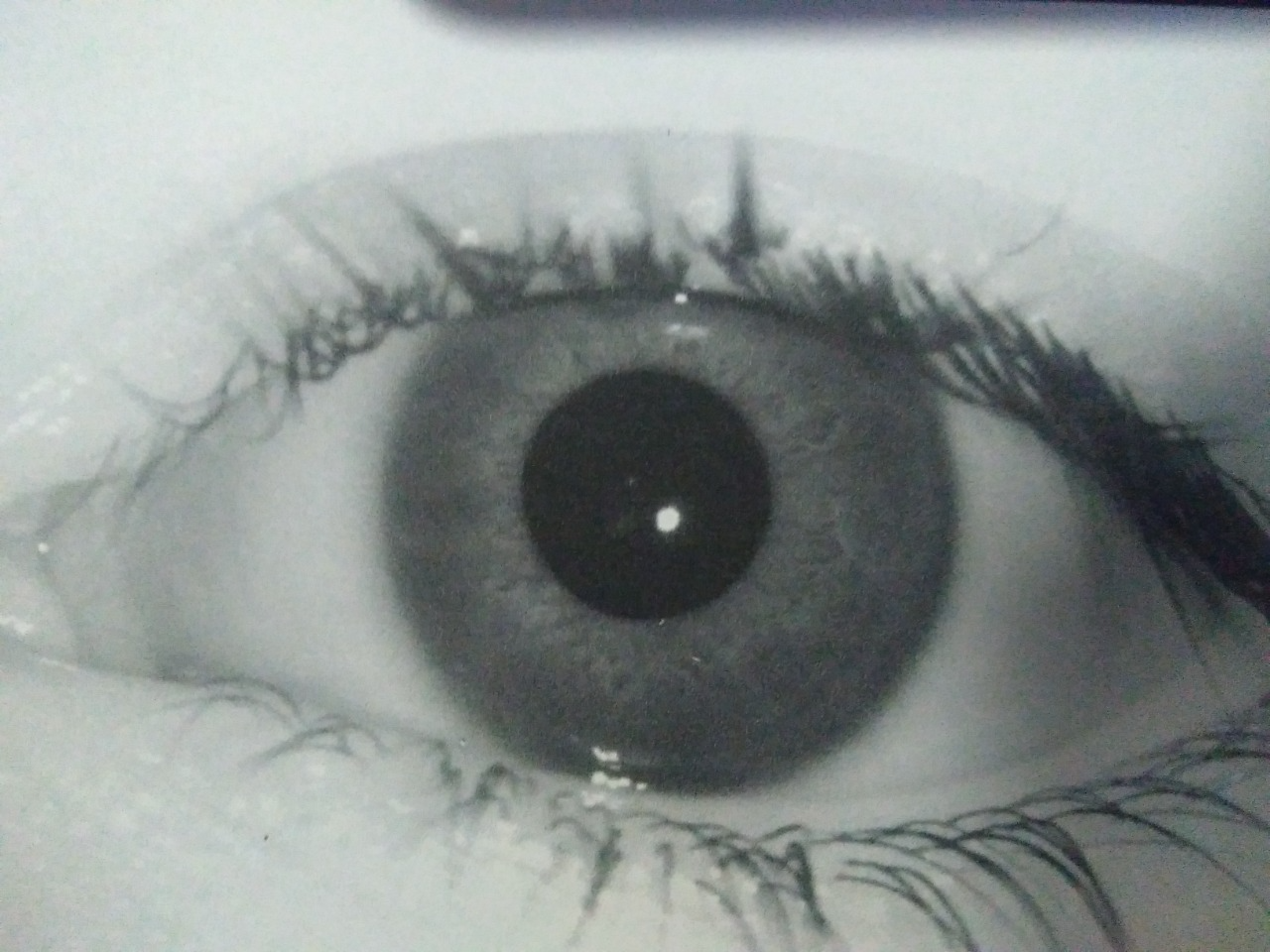}}
    \hfill
    \subcaptionbox{\centering Print-out Nokia-9-PureView}{\includegraphics[scale=0.09]{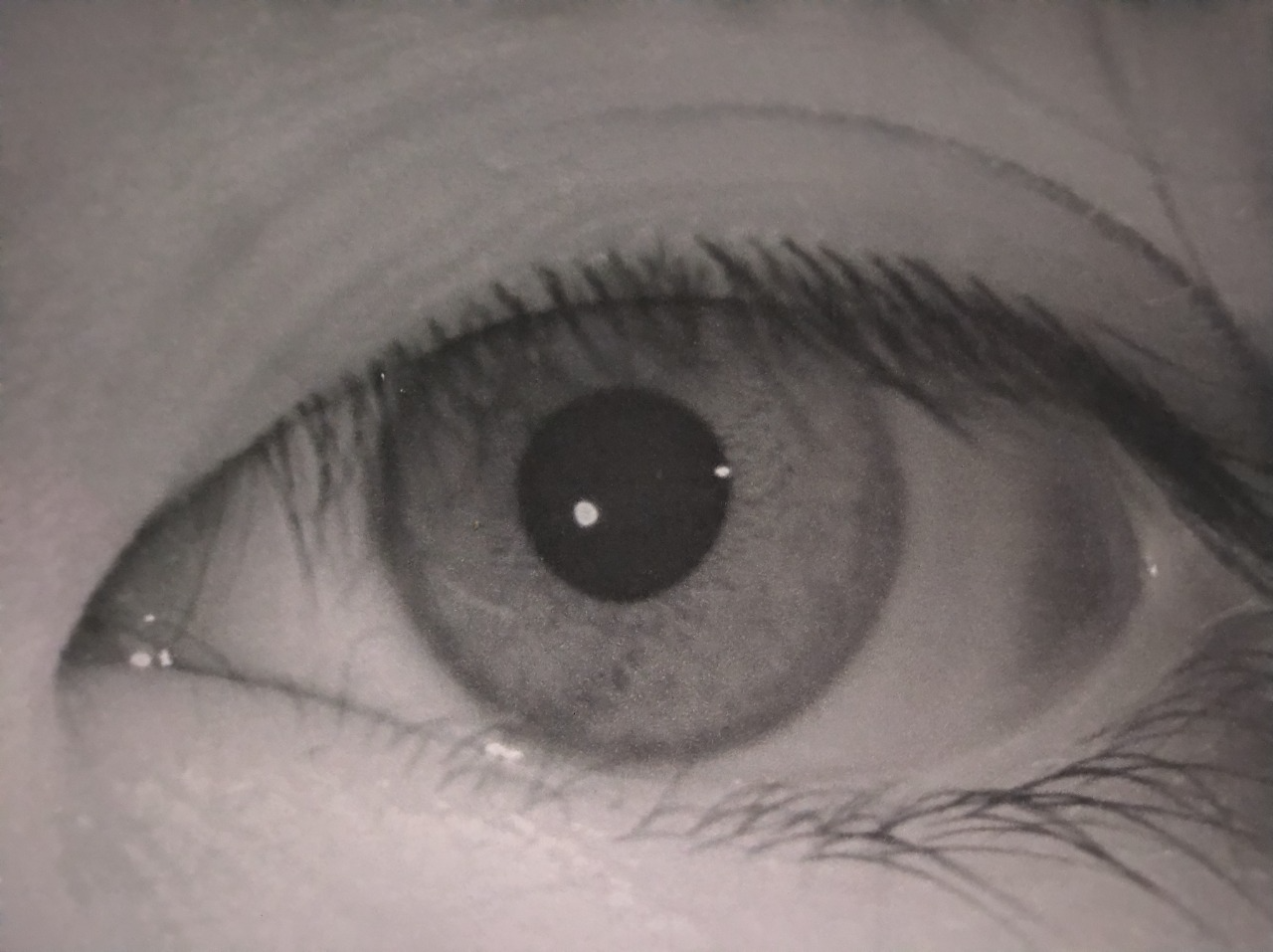}}
    \hfill
    \subcaptionbox{\centering Patterned eye}{\includegraphics[scale=0.19]{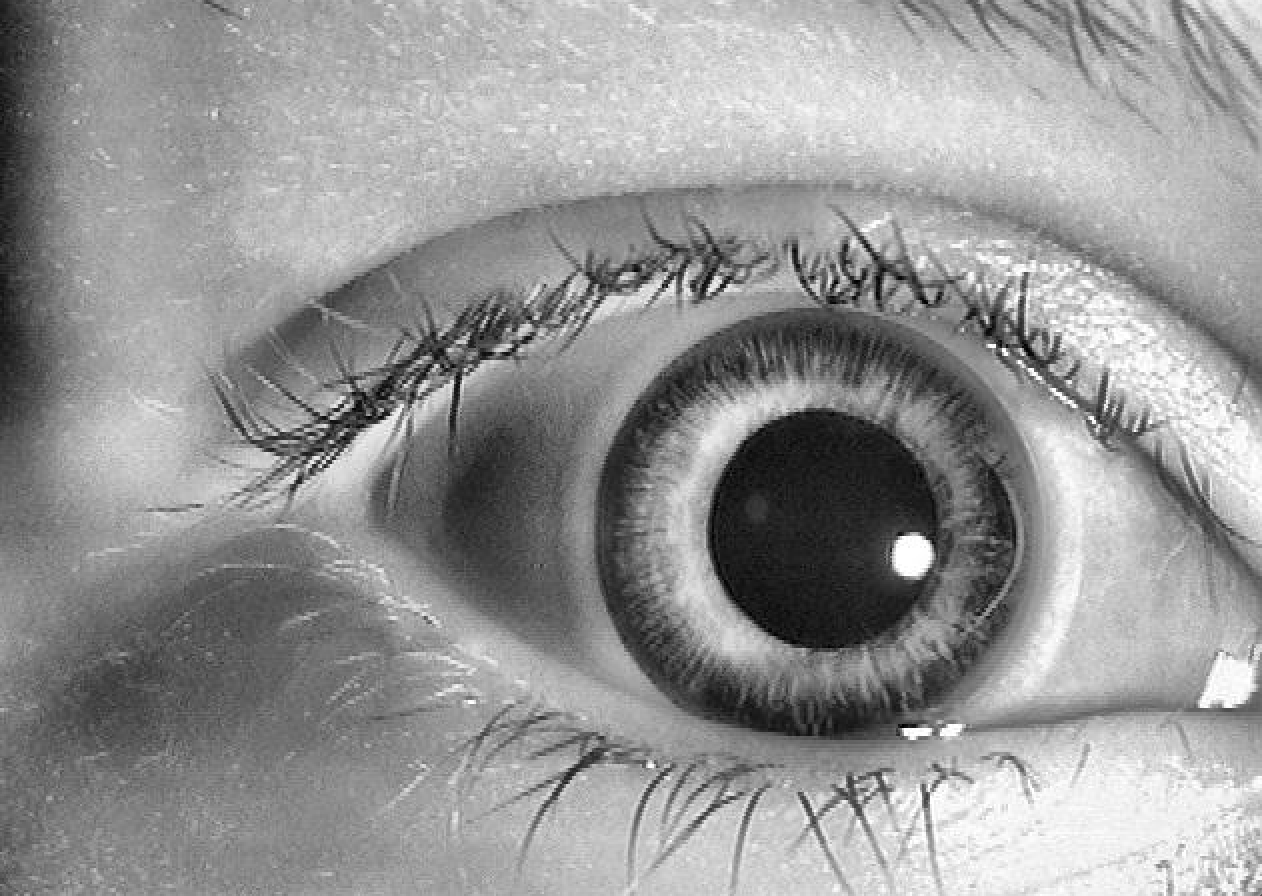}}
\caption{Example images of all presentation attack instruments in the database.}
\label{fig:Example_images}
\end{figure*}

\section{Method}
\label{sec:method}

This work explored several variations of the DinoV2 self-supervised vision transformers' backbones, such as DinoV2-ViTB14, DinoV2-ViTS14 and DinoV2-ViTL. Each one of these models contains 86.9, 22.2 and 304.0 Million parameters, respectively. 

The DinoV2 architecture consists of a backbone common to many computer vision tasks, with a task-specific head connected to the end of the backbone. For instance, when performing image classification, the head includes a linear classification layer that takes the vector representation produced by the backbone and uses it to categorize the image into different classes. In summary, DinoV2 features a general-purpose backbone for computer vision tasks, while the linear head could be added to tailor a specific task at hand. Further on that, based on the same fine-tuning technique, we also explored VisualOpenClip models such as VIT-B-32 with 88.0, VIT-B-16 with 86.3 and ViT-L-14 with 304 Million parameters, respectively. In this case, when we only use the visual approach task and remove the text prediction. This branch is called VisualOpenClip. 

\subsection{Database}

This study used the same dataset as the LivDet-Iris 2020 competition. Additionally, we incorporated three supplementary databases of iris images \cite{Hybrid-tapia}. 

The first database, named "Iris-CL1," contains near-infrared (NIR) bona fide images captured using an Iritech iris sensor, with a resolution of $640\times480$ pixels. The second database, called "iris-printed-CL1," consists of high-quality presentation attack images of printed presentation attack images. This database was created to enhance the challenge posed by printed iris species, as the visible patterns in the printed images from the LivDet-Iris 2020 database make them relatively easy to differentiate from bona fide images. The iris-printed-CL1 database contains 1,800 images captured using two smartphone devices, with 900 images from each device: the Nokia-9-PureView, which has an image resolution of $1280\times957$ pixels, and the Motorola Moto-G4-Play, with an image resolution of $1280\times960$ pixels. Only the red channel of the images was used for analysis. Figure~\ref{fig:Example_images} show examples of images of printed iris species. 

The third database is the Warsaw-BioBase-Post-Mortem-Iris v3.0  database~\cite{TROKIELEWICZ2020103866}. This database contains a total of 1,094 NIR images and 785 visible-light images (obtained with Olympus-TG-3) collected from 42 post-mortem subjects. Table~\ref{tab:db} shows a summary of all datasets available from LivDet-Iris 2020, plus Cadaver images, iris-CL1, and iris-printed-CL2. The new total images available is 27,964. %This is more than two times the number of pictures shown in Table~\ref{tab:db}

\begin{table}[]
\centering
\caption{Summary PAD iris dataset.}
\label{tab:db}
\resizebox{\columnwidth}{!}{%
\begin{tabular}{|cc|r|r|r|r|c|}
\hline
\multicolumn{1}{|c|}{\textbf{Class}} &
  \textbf{PAI species} &
  \multicolumn{1}{c|}{\textbf{Train}} &
  \multicolumn{1}{c|}{\textbf{Val}} &
  \multicolumn{1}{c|}{\textbf{Test}} &
  \multicolumn{1}{c|}{\textbf{Num Im.}} &
  \textbf{Sensors} \\ \hline
\multicolumn{1}{|c|}{\begin{tabular}[c]{@{}c@{}}Bona\\ Fide\end{tabular}} &
  --- &
  6,694 &
  1,062 &
  5,773 &
  13,530 &
  \begin{tabular}[c]{@{}c@{}}LG4000\\ AD/TD 100\\ iCam 700\end{tabular} \\ \hline
\multicolumn{1}{|c|}{Attack} &
  Cadaver &
  448 &
  531 &
  754 &
  1,773 &
  \begin{tabular}[c]{@{}c@{}}IriTech\\ IriShield\end{tabular} \\ \hline
\multicolumn{1}{|c|}{Attack} &
  \begin{tabular}[c]{@{}c@{}}Cont. Lenses\\ Textured\end{tabular} &
  3,583 &
  900 &
  3,244 &
  7,727 &
  \begin{tabular}[c]{@{}c@{}}LG4000\\ AD/TD 100\\ iCam 700\\ MotoG4\\ Gplay\end{tabular} \\ \hline
\multicolumn{1}{|c|}{Attack} &
  \begin{tabular}[c]{@{}c@{}}Printed\\ Prosthetic\\ Display\end{tabular} &
  4,090 &
  1,896 &
  2,305 &
  8,291 &
  \begin{tabular}[c]{@{}c@{}}Iris ID\\ iCAM700\end{tabular} \\ \hline
\multicolumn{2}{|c|}{\textbf{Total}} &
  \textbf{11,810} &
  \textbf{4,384} &
  \textbf{11,770} &
  \textbf{27,964} &
   \\ \hline
\end{tabular}%
}
\end{table}

\subsection{Metrics}
\label{sec:metric}

The ISO/IEC 30107-3 standard\footnote{\url{https://www.iso.org/standard/67381.html}} presents methodologies for evaluating the performance of PAD algorithms for biometric systems. The APCER metric measures the proportion of attack presentations---for each different PAI---incorrectly classified as bona fide presentations. This metric is calculated for each PAI, where the worst-case scenario is considered. Equation~\ref{eq:apcer} details how to compute the APCER metric, in which the value of $N_{PAIS}$ corresponds to the number of attack presentation images, where $RES_{i}$ for the $i$th image is $1$ if the algorithm classifies it as an attack presentation, or $0$ if it is classified as a bona fide presentation (real image).

\begin{equation}\label{eq:apcer}
    {APCER_{PAIS}}=1 - (\frac{1}{N_{PAIS}})\sum_{i=1}^{N_{PAIS}}RES_{i}
\end{equation}

Additionally, the BPCER metric measures the proportion of bona fide presentations mistakenly classified as attack presentations or the ratio between false rejection to total bona fide attempts. The BPCER metric is formulated according to equation~\ref{eq:bpcer}, where $N_{BF}$ corresponds to the number of bona fide presentation images, and $RES_{i}$ takes identical values of those of the APCER metric.

\begin{equation}\label{eq:bpcer}
    BPCER=\frac{\sum_{i=1}^{N_{BF}}RES_{i}}{N_{BF}}
\end{equation}

These metrics effectively measure to what degree the algorithm confuses presentations of attack images with bona fide images and vice versa. The APCER and BPCER metrics depend on a decision threshold.

\section{Experiment and Results}
\label{sec:expandresults}

In order to evaluate the performance of state-of-the-art deep learning models in a fine-tuning task and trained from scratch, we implemented the following binary models (bona fide and attack) to compare the performance with the foundation models. The CNN models were MobileNetV3\_large \cite{mbv3}, ResNet34 \cite{resnet}, ResNet101 \cite{resnet}, EfficientNetB0-V2 \cite{efficientnet}, DenseNet121 \cite{DenseNet} and SwinTranformerV2 \cite{swin}. For VisualOpenClip, the pre-trained backbone "laion400m\_e32" was used.

\begin{figure*}[]
      \centering
      \includegraphics[scale=0.5]{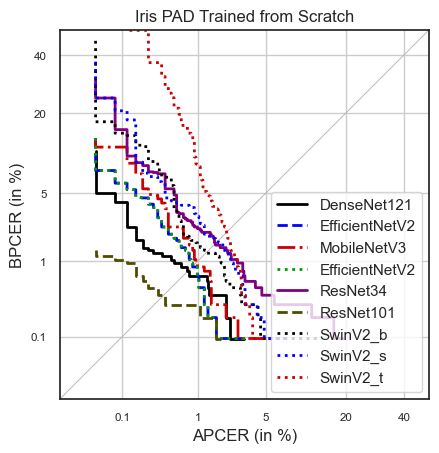}
      \includegraphics[scale=0.5]{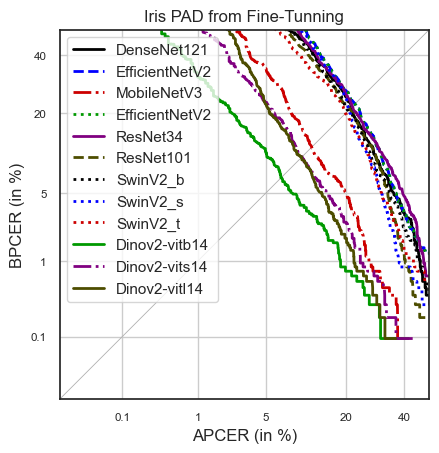}
      \includegraphics[scale=0.5]{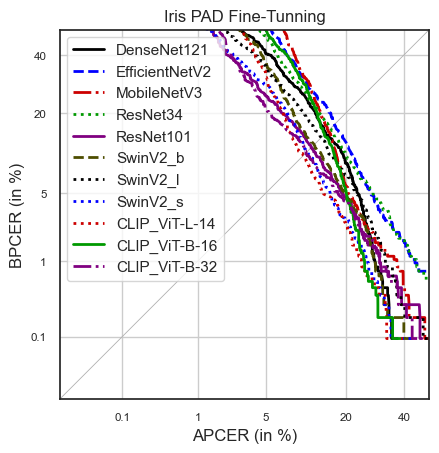}
      \caption{Left: Iris PAD results trained from scratch. Middle: Iris PAD results trained from fine-tuning on DinoV2. Right: Iris PAD results trained from fine-tuning on VisualOpenClip.}
      \label{fig:det}
\end{figure*}

\subsection{Experiment 1 - Scratch}

Several networks were used to train a PAD model from scratch to compare the results of the fine-tuning approach considering foundation models. DinoV2 and VisualOpenClip were not used in this experiment because we did not have any capabilities to train this model from scratch. All the images were resized to $224 \times 224$ to adapt to the image resolution initially used by DinoV2 and CLIP during training with data augmentation of a random crop of $480\times360$, jitter illuminations, a small rotation of 5 degrees and a horizontal flip.

\subsection{Experiment 2 - Fine-Tuning}
This experiment explores the results of DinoV2 and VisualOpenClip in a fine-tuning task, which means they do not include any NIR iris bona fide images or attacks in the original trained set. Because of that, we add a small head to adjust the weight of the linear classifier. All traditional deep learning methods fail or obtain a lower performance because they do not have generalization capabilities to other domains. In this experiment, all the layers except the head are frozen, and the feature extracted vector was sent to a neural network classifier with two layers and a ReLU activation function. Binary cross-entropy was used with grid search to find the best learning rate. Values from $1e-3$ , $1e-4$, $1e-5$ and $1e-6$ was explored. The same pre-processing operation from Experiment 1 was applied related to data augmentation and grid search parameters. Overall, the ResNet101 model reached lower performances.

Figure \ref{fig:det} shows the DET curves for state-of-the-art deep learning models applied in a fine-tuning task in comparison with DinoV2 and VisualOpenClip. In Figure 2 (Middle), the green line shows that the DinoV2-ViTB14 model outperforms all the methods based on deep learning trained using the ImageNet weights. 

Table \ref{tab:fewshot} and Table \ref{tab:scratch} present the results of the Equal Error Rate (EER) and three different operational points: BPCER10, BPCER20, and an additional metric BPCER100, for both fine-tuning and trained-from-scratch models. The best results are highlighted in bold; notably, DinoV2-ViTB14 achieved an EER of 6.77\% and a BPCER10 of 3.38\%, surpassing all other models. The VisualOpenClip-ViT-L14 also obtained one of the best results compared to traditional methods.
This performance is particularly impressive considering that this model was not trained with NIR iris images, demonstrating robust prediction capabilities that are sufficient for developing a competitive Presentation Attack Detection (PAD) system without restrictions related to privacy-sensitive images representing bona fide attacks during training.

\begin{table}[]
\centering
\caption{Fine-tuning for Iris Presentation Attack Detection. All the results are in (\%).}
\label{tab:fewshot}
\begin{tabular}{|c|c|c|c|c|}
\hline
\begin{tabular}[c]{@{}c@{}}Model \\ Feature\\ Extractor\end{tabular} &
  \begin{tabular}[c]{@{}c@{}}EER \\ (\%)\end{tabular} &
  \begin{tabular}[c]{@{}c@{}}BPCER10 \\ (\%)\end{tabular} &
  \begin{tabular}[c]{@{}c@{}}BPCER20 \\ (\%)\end{tabular} &
  \begin{tabular}[c]{@{}c@{}}BPCER100 \\ (\%)\end{tabular} \\ \hline
\multicolumn{1}{|l|}{CLIP-ViT-B-16} &
  \multicolumn{1}{l|}{14.432} &
  \multicolumn{1}{l|}{30.602} &
  \multicolumn{1}{l|}{40.951} &
  \multicolumn{1}{l|}{76.836} \\ \hline
CLIP-ViT-B-32 & 10.741          & 21.374          & 60.075          & 77.118          \\ \hline
CLIP-ViT-L-14 & \textbf{10.308} & \textbf{10.640} & \textbf{32.39}  & \textbf{74.950} \\ \hline
DenseNet121   & 21.694          & 46.139          & 57.156          & 77.118          \\ \hline
Dinov2-vitb14 & \textbf{6.775}  & \textbf{3.860}  & \textbf{10.828} & \textbf{31.732} \\ \hline
Dinov2-vitl14 & 9.923           & 9.510           & 22.598          & 63.747          \\ \hline
Dinov2-vits14 & 9.970           & 9.981           & 23.258          & 54.048          \\ \hline
Efficientv2-S & 22.704          & 51.600          & 71.374          & 93.596          \\ \hline
MobileNetv3   & \textbf{11.810} & \textbf{16.195} & \textbf{34.745} & \textbf{71.845} \\ \hline
ResNet34      & 21.954          & 46.986          & 66.007          & 88.323          \\ \hline
ResNet101     & 21.223          & 42.749          & 58.474          & 84.745          \\ \hline
Swinv2-b      & 20.665          & 45.291          & 59.416          & 78.719          \\ \hline
Swinv2-s      & 20.904          & 49.058          & 69.774          & 84.839          \\ \hline
Swinv2-t      & 19.961          & 38.323          & 53.672          & 75.800          \\ \hline
\end{tabular}
\end{table}

In the case of no restriction of images regarding the number of subjects and attacks trained from scratch, it still shows impressive results. In this case, DenseNet121 outperform the results of foundation models.

It is important to emphasize that the inference of all images is conducted independently, without utilizing any cloud services that could expose the datasets.

\begin{table}[]
\centering
\caption{Models trained from scratch for Iris Presentation Attack Detection. All the results are in (\%).}
\label{tab:scratch}
\begin{tabular}{|c|c|c|c|c|}
\hline
\begin{tabular}[c]{@{}c@{}}Model \\ Feature \\ Extractor\end{tabular} &
  \begin{tabular}[c]{@{}c@{}}EER \\ (\%)\end{tabular} &
  \begin{tabular}[c]{@{}c@{}}BPCER10 \\ (\%)\end{tabular} &
  \begin{tabular}[c]{@{}c@{}}BPCER20 \\ (\%)\end{tabular} &
  \begin{tabular}[c]{@{}c@{}}BPCER100 \\ (\%)\end{tabular} \\ \hline
DenseNet121   & \textbf{0.750} & \textbf{0.0} & \textbf{0.0} & \textbf{0.659} \\ \hline
Efficientv2-S & 0.851          & 0.0          & 0.0          & 0.65           \\ \hline
MobileNetv3   & 0.943          & 0.0          & 0.094        & 0.847          \\ \hline
ResNet34      & 1.540          & 0.282        & 0.376        & 2.259          \\ \hline
ResNet101     & 0.365          & 0.0          & 0.0          & 0.282          \\ \hline
Swinv2-b      & 1.241          & 0.094        & 0.094        & 1.506          \\ \hline
Swinv2-s      & 1.646          & 0.0          & 0.094        & 3.013          \\ \hline
Swinv2-t      & 2.012          & 0.0          & 0.0          & 8.757          \\ \hline
\end{tabular}
\end{table}

\section{Conclusion}
\label{sec:conclusions}

The results of this approach show that the foundation models based on DinoV2-ViTB14 and VisualOpenClip are excellent options for creating an iris PAD system. This option opens a new night in biometric tasks and is particularly relevant in situations where there is a scarcity of data, which can often pose significant challenges.

%not a large number of images are available to train a method from scratch.

%%%%%%%%%%%%%%%%%%%%%%%%%%%%%%%%%%%%%%%%%%%%%%%%%%%%%%%%%%%%%%%%%%%%%%%%%%%%%%%%
%\section{ACKNOWLEDGMENTS}
%Removed for blind revision.

%%%%%%%%%%%%%%%%%%%%%%%%%%%%%%%%%%%%%%%%%%%%%%%%%%%%%%%%%%%%%%%%%%%%%%%%%%%%%%%%

%%%%%%%%%%%%%%%%%%%%%%%%%%%%%%%%%%%%%%%%%%%%%%%%%%%%%%%%%%%%%%%%%%%%%%%%%%%%%%%%
%\newpage
%\section*{ETHICAL IMPACT STATEMENT}
%This research adheres to all ethical guidelines established by Face and Gesture 2025. The dataset was obtained in compliance with the providers' recommendations. All data has been anonymized to ensure that no individual can be discriminated against based on gender, ethnicity, or any other characteristic. Furthermore, all datasets have previously been utilized in open NIR iris presentation attack competitions.
%%%%%%%%%%%%%%%%%%%%%%%%%%%%%%%%%%%%%%%%%%%%%%%%%%%%%%%%%%%%%%%%%%%%%%%%%%%%%%%%

{\small
\bibliographystyle{ieee}
\bibliography{egbib}
}

\end{document}